\documentclass[sigconf,natbib=true,anonymous=false]{acmart}

\AtBeginDocument{%
  }

\copyrightyear{2024}
\acmYear{2024}
\setcopyright{acmlicensed}\acmConference[CIKM '24]{Proceedings of the 33rd ACM International Conference on Information and Knowledge Management}{October 21--25, 2024}{Boise, ID, USA}
\acmBooktitle{Proceedings of the 33rd ACM International Conference on Information and Knowledge Management (CIKM '24), October 21--25, 2024, Boise, ID, USA}
\acmDOI{10.1145/3627673.3679920}
\acmISBN{979-8-4007-0436-9/24/10}

\usepackage{subfigure}
\usepackage{multirow}
\usepackage{mathtools}
\usepackage{color,colortbl}

\definecolor{Gray}{gray}{0.9}

\newcolumntype{x}{>{\columncolor{Gray}}l}
\newcolumntype{y}{>{\columncolor{Gray}}l}

\begin{document}

\title{Exploiting Preferences in Loss Functions for Sequential Recommendation via Weak Transitivity}

\author{Hyunsoo Chung}
\affiliation{%
  \institution{Omnious AI}
  \city{Seoul}
  \country{South Korea}
}
\email{hyunsoo.chung@omnious.com}

\author{Jungtaek Kim}
\affiliation{%
  \institution{University of Pittsburgh}
  \city{Pittsburgh}
  \state{Pennsylvania}
  \country{United States}
}
\email{jungtaek.kim@pitt.edu}

\author{Hyungeun Jo}
\affiliation{%
  \institution{Omnious AI}
  \city{Seoul}
  \country{South Korea}
}
\email{hyungeun.jo@omnious.com}

\author{Hyungwon Choi}
\affiliation{%
  \institution{Omnious AI}
  \city{Seoul}
  \country{South Korea}
}
\email{hyungwon.choi@omnious.com}

\begin{abstract}
A choice of optimization objective is immensely pivotal in the design of a recommender system as it affects the general modeling process of a user's intent from previous interactions. 
Existing approaches mainly adhere to three categories of loss functions: pairwise, pointwise, and setwise loss functions.
Despite their effectiveness, a critical and common drawback of such objectives is viewing the next observed item as a unique positive while considering all remaining items equally negative.
Such a binary label assignment is generally limited to assuring a higher recommendation score of the positive item, neglecting potential structures induced by varying preferences between other unobserved items.
To alleviate this issue, we propose a novel method that extends original objectives to explicitly leverage the different levels of preferences as relative orders between their scores.
Finally, we demonstrate the superior performance of our method compared to baseline objectives.
\end{abstract}




\maketitle

\begin{figure}[t]
  \centering
  \includegraphics[width=0.99\linewidth]{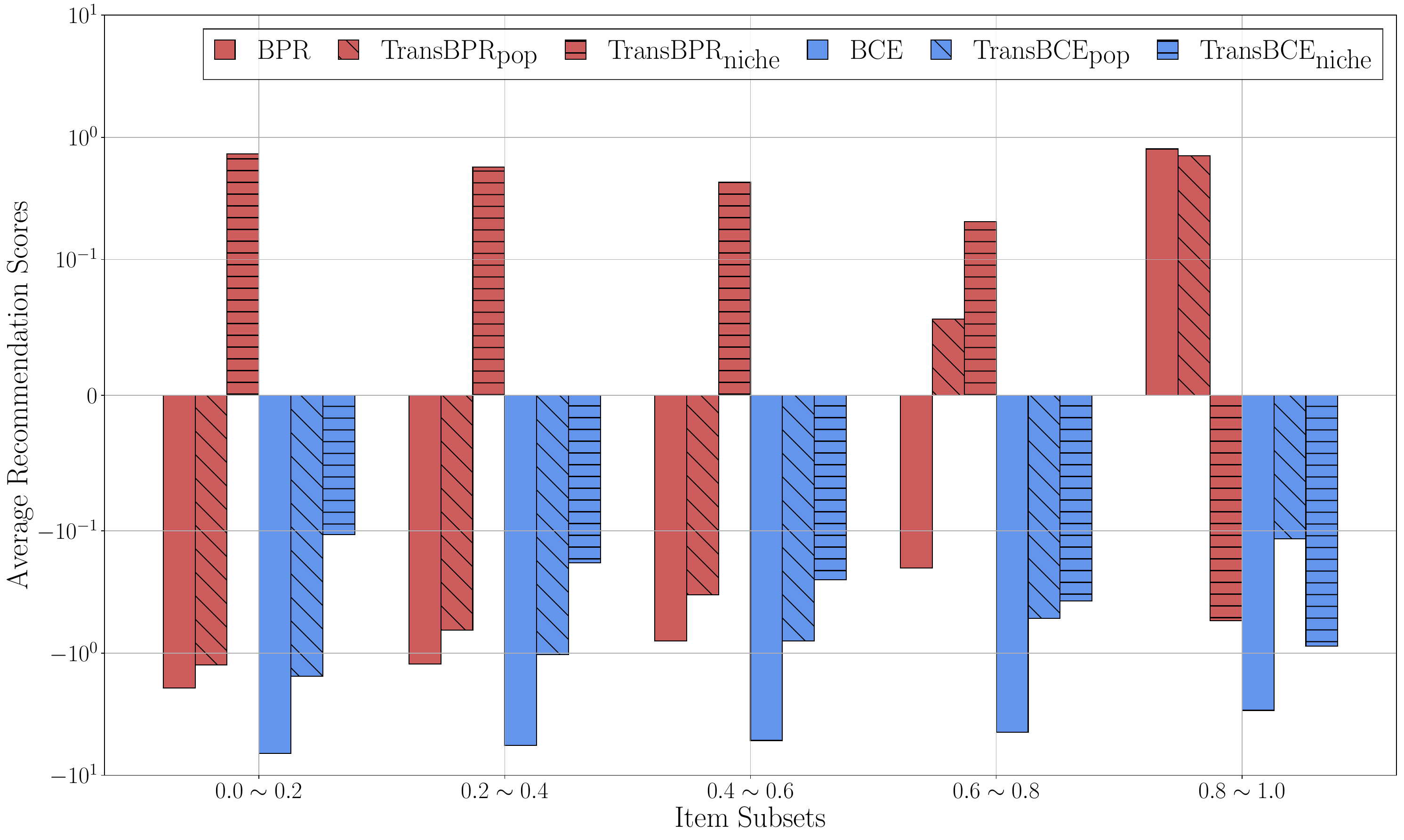}
  \caption{Average recommendation scores of item subsets on the Amazon Beauty dataset trained with BPR (red solid), BCE (blue solid), and our transitive extensions (hatch).
  Each item subset in $x$-axis equally consists of $20\%$ of the total items divided based on their popularity.
  Bars with diagonal hatch assume that a user generally prefer popular items whereas bars with horizontal hatch regard niche items more preferred.}
  \Description{Average recommendation scores of item subsets}
  \label{fig:teaser}
\end{figure}

\section{Introduction}
\label{sec:introduction}

Distinguishing items by relevance to an each user's previous interaction history is the essence of most applied sequential recommender systems.
Conventional approaches~\cite{hidasi2015session, tang2018personalized, wu2020sse, sun2019bert4rec} frame this task as learning the feature representations of a chronologically-ordered list of user's interacted items and a candidate item.
By computing the inner product between two vectors, we then obtain a recommendation score that quantitatively indicates the relevance.
On that basis, a prevalent training process reformulates the problem into a supervised learning framework with binary labels, where the next interacted item of each user becomes a unique positive to its sequence of previous items.
Such a procedure often accompanies negative sampling~\cite{chen2017sampling, rendle2014improving, ding2020simplify, lian2020personalized, chen2022generating, chen2023fairly} among the rest of items since the score computation of all items during each training step inevitably results in severe inefficiency~\cite{zhou2021contrastive, klenitskiy2023turning}.

In this line of research, numerous approaches have adopted advanced neural networks as feature encoders~\cite{Zhou2020S3Rec, fan2021continuous, fan2022sequential, 
kang2018self, zhou2022filter, wang2019neural, he2020lightgcn, wu2021self, liu2021interest, chen2023heterogeneous} to capture more complex correlations.
Apart from vast architectural improvements, the majority of models yet utilize one of three types of loss functions as an optimization objective:
pairwise, e.g., BPR~\cite{Rendle2009BPR},
pointwise, e.g., BCE~\cite{chang2021sequential, zhou2021temporal},
and setwise, e.g., SSM~\cite{wu2024effectiveness} functions.
While the formulation of each function varies, they all share the desired result of increasing the score of the unique positive than the scores of the other items.
Notably, most recent studies on loss functions and negative sampling strategies~\cite{wang2017irgan, zhang2013optimizing, wang2018incorporating, lobato2014probabilistic, zhang2019nscaching, ding2020simplify, chen2023fairly} aim for improved training efficiency and robustness to false negatives within three core objectives.
Despite their effectiveness, the model trained with such objectives consequently learns to regard unobserved items as equally negative with their labels simply set to zero.
In real-world recommendation, however, a subset of negative items is occasionally more favorable than the other due to various side factors, e.g., item popularity.
Unfortunately, the current scheme of binary label assignments hinders from fully taking advantage of diverse levels of preferences among unobserved items.

To address the aforementioned issue, we first derive an inductive relation, dubbed weak transitivity, which represents preference-driven orders of item scores.
We then propose novel extensions of original loss functions that directly leverage this weak transitivity in their forms.
Consequently, the recommendation scores of unobserved items are aligned with respect to their preferences.
Figure~\ref{fig:teaser} highlights such a property of resulting recommendation policy trained with our proposed family of objectives.
Opposed to the results of BPR~\cite{Rendle2009BPR} and BCE~\cite{kang2018self}, item scores from our transitive extensions $\textrm{TransBPR}_{\textrm{pop}}$ and $\textrm{TransBCE}_{\textrm{pop}}$ (horizontal hatch) are generally proportional to their popularity.
Meanwhile, scores from $\textrm{TransBPR}_{\textrm{niche}}$ and $\textrm{TransBCE}_{\textrm{niche}}$ (diagonal hatch) are inversely proportional to their popularity, favoring more niche (i.e., unique) items.
To cap it all, we serve the predefined preferences of items as an additional supervisory factor for their recommendation scores.

It is noteworthy that in this work we do not propose any new distributions for negative sampling but instead introduce the modification of original objectives.
Hence, we solely utilize a combination of item popularity and uniform distributions for negative sampling, easily accessible in implicit settings.
We validate the effectiveness of our proposed extensions compared to the original loss functions and their renowned variants on four sequential recommendation benchmarks.
In all settings, our approaches substantially improve the recommendation performance compared to baseline methods.

\section{Background}

In this section, we describe a sequential recommendation task and the details of representative loss functions for recommendation.

\subsection{Problem Statement}

Let $\mathcal{U} = \{u_1, \ldots, u_M\}$ and $\mathcal{I} = \{i_1, \ldots, i_N\}$ denote a set of users and items where $M$ and $N$ are maximum numbers of users and items, respectively.
Given a chronologically ordered history $h_u = \{i_1^u, \ldots, i_t^u\}$ of observed items for a user $u$,
a goal of sequential recommendation is to recommend the most relevant next item $i_{t+1}^u$.
We first embed the history $h_u$ and a candidate item $i$ onto vectors $h'_u$ and $i'$, respectively.
Accordingly, a recommendation score $\hat{s}_{ui}$ is calculated via an inner product between two embeddings.

\subsection{Training Objectives}

\paragraph{Pairwise Objective} 
Bayesian personalized ranking (BPR)~\cite{Rendle2009BPR} models personalized ranking of items.
It forces the score of the positive (i.e., next interacted) item to be higher than the scores of the rest unobserved items.
With a dataset $\mathcal{D}_s$ composed of a triplet $(u,i,j)$ of which item $i$ as a positive and item $j$ as a sampled negative to a user $u$,
the corresponding loss is formulated as follows:
\begin{equation}
    \mathcal{L}_{\textrm{BPR}} = -
    \sum_{(u,i,j) \in \mathcal{D}_s}
    \log\sigma(\hat{s}_{ui} - \hat{s}_{uj}),
    \label{eqn:bpr_loss}
\end{equation}
where $\sigma(\cdot)$ is a sigmoid function.
A resulting term $\sigma(\hat{s}_{ui} - \hat{s}_{uj})$ is the probability of a user $u$ preferring an item $i$ more than an item $j$.

\paragraph{Pointwise Objective}
Similar to the pairwise loss form, binary cross-entropy (BCE)~\cite{kang2018self} casts the recommendation problem into binary classification with a single sampled negative.
Hence with a dataset $\mathcal{D}_s$ of a triplet $(u,i,j)$, the loss is defined as below:
\begin{equation}
    \mathcal{L}_{\textrm{BCE}} = -
    \sum_{(u,i,j) \in \mathcal{D}_s}
    \log(\sigma(\hat{s}_{ui})) + 
    \log(1 - \sigma(\hat{s}_{uj})).
    \label{eqn:bce_loss}
\end{equation}

\paragraph{Setwise Objective}
The sampled softmax loss (SSM)~\cite{wu2024effectiveness} turns the problem into a multi-class classification with a finite number of sampled negatives.
With a dataset $\mathcal{D}_m$ of a triplet $(u,i, \mathcal{N}_j)$ consisting of the positive $i$ and a set of multiple negative samples $\mathcal{N}_j$ to the user $u$, the loss optimizes the probability of the positive as:
\begin{equation}
    \mathcal{L}_{\textrm{SSM}} = -
    \sum_{(u,i,\mathcal{N}_j) \in \mathcal{D}_m}
    \log\frac{\exp(\hat{s}_{ui})}
    {\exp(\hat{s}_{ui}) + 
    \sum_{j \in \mathcal{N}_j}\exp(\hat{s}_{uj})}.
    \label{eqn:ssm_loss}
\end{equation}

\begin{table}[t]
    \caption{Statistics of four preprocessed datasets.}
    \label{tab:datasets}
    \setlength{\tabcolsep}{6pt}
    \begin{center}
    \begin{tabular}{lrrrr}
    \toprule
    \textbf{Dataset} & \textbf{\#Interactions} & \textbf{\#Users} & \textbf{\#Items} &  \textbf{Density}\\
    \midrule
    Beauty & 198,502 & 22,363 & 12,101 & 0.00073\\
    Toys & 167,597 & 19,412 & 11,924 & 0.00072\\
    Sports & 296,337 & 35,598 & 18,357 & 0.00045\\
    Yelp & 317,182 & 30,499 & 20,068 & 0.00052\\
    \bottomrule
    \end{tabular}
    \end{center}
\end{table}

\begin{table*}[t]
    \caption{Quantitative results of different methods on public datasets in terms of HR and NDCG. 
    The number of recommended items is fixed to 10. 
    Reported metrics in bold are best performing methods whereas underlined numbers are second to the best.}
    \label{tab:main_results}
    \setlength{\tabcolsep}{10pt}
    \begin{center}
    \begin{tabular}{lxxyxyxyxy}
    \toprule
    \rowcolor{white}
     &
     &
    \multicolumn{2}{c}{\textbf{Beauty}} &
    \multicolumn{2}{c}{\textbf{Toys}} &
    \multicolumn{2}{c}{\textbf{Sports}} &
    \multicolumn{2}{c}{\textbf{Yelp}} \\
    \rowcolor{white}
    \multirow{-2}{*}{\textbf{Type}} & 
    \multirow{-2}{*}{\textbf{Loss}} & 
    HR & NDCG & HR & NDCG & 
    HR & NDCG & HR & NDCG \\
    \midrule
    \rowcolor{white}
    & $\textrm{BPR}$ &
    0.0470 & 0.0213 & 0.0518 & 0.0244 & 
    0.0263 & 0.0122 & 0.0555 & 0.0328 \\
    \rowcolor{white}
    & $\textrm{GBPR}$ &
    0.0529 & 0.0246 & 0.0618 & 0.0281 &
    0.0277 & 0.0121 & \textbf{0.0607} & 0.0346 \\
    & $\textrm{TransBPR}_{\textrm{pop}}$ &
    \textbf{0.0674} & \textbf{0.0293} & 
    \underline{0.0747} & \underline{0.0338} & 
    \textbf{0.0397} & \textbf{0.0179} & 
    \underline{0.0576} & \textbf{0.0362} \\
    \multirow{-4}{*}{Pairwise} & 
    $\textrm{TransBPR}_{\textrm{niche}}$ & 
    \underline{0.0651} & \underline{0.0289} & 
    \textbf{0.0777} & \textbf{0.0358} & 
    \underline{0.0372} & \underline{0.0171} & 
    0.0566 & \underline{0.0355} \\
    \midrule
    \rowcolor{white}
    & $\textrm{BCE}$ &
    0.0511 & 0.0229 & 0.0590 & 0.0274 & 
    0.0290 & 0.0130 & 0.0529 & 0.0306 \\
    \rowcolor{white}
    & $\textrm{gBCE}_{1}$ &
    0.0545 & 0.0245 & 0.0656 & 0.0297 &
    0.0294 & 0.0135 & 0.0551 & 0.0319 \\
    & $\textrm{TransBCE}_{\textrm{pop}}$ & 
    \textbf{0.0730} & \textbf{0.0324} & 
    \textbf{0.0805} & \textbf{0.0372} &
    \textbf{0.0413} & \textbf{0.0186} & 
    \underline{0.0561} & \textbf{0.0360} \\
    \multirow{-4}{*}{Pointwise} & 
    $\textrm{TransBCE}_{\textrm{niche}}$ &
    \underline{0.0720} & \underline{0.0322} & 
    \underline{0.0800} & \underline{0.0369} & 
    \underline{0.0399} & \underline{0.0179} & 
    \textbf{0.0569} & \underline{0.0358} \\
    \midrule
    \rowcolor{white}
    & $\textrm{SSM}$ &
    0.0656 & 0.0318 & 0.0673 & 0.0351 &
    0.0381 & 0.0185 & 0.0616 & 0.0350 \\
    \rowcolor{white}
    & $\textrm{InfoNCE}$ &
    0.0632 & 0.0318 & 0.0742 & 0.0395 &
    0.0363 & 0.0188 & 0.0483 & 0.0255 \\
    \rowcolor{white}
    & $\textrm{BPR}\textrm{-}\textrm{DNS}$ &
    \underline{0.0776} & 0.0356 & 0.0839 & 0.0386 &
    0.0406 & 0.0187 & 0.0523 & 0.0349 \\
    \rowcolor{white}
    & $\textrm{gBCE}_{\mathcal{N}}$ &
    0.0725 & 0.0362 & 0.0783 & 0.0407 &
    \underline{0.0415} & \underline{0.0205} & 
    0.0632 & 0.0359 \\
    & $\textrm{TransSSM}_{\textrm{pop}}$ & 
    \textbf{0.0843} & \textbf{0.0395} & 
    \textbf{0.0922} & \textbf{0.0439} &
    \textbf{0.0499} & \textbf{0.0229} & 
    \textbf{0.0693} & \textbf{0.0418} \\
    \multirow{-6}{*}{Setwise} & 
    $\textrm{TransSSM}_{\textrm{niche}}$ & 
    0.0774 & \underline{0.0370} & 
    \underline{0.0874} & \underline{0.0417} & 
    0.0399 & 0.0187 & 
    \underline{0.0642} & \underline{0.0385} \\
    \bottomrule
    \end{tabular}
    \end{center}
\end{table*}

\section{Our Method}

Here, we introduce weak transitivity between unobserved items,
and then describe how to integrate it to original objectives.

\subsection{Weak Transitivity}

Inducing orders on recommendation scores of unobserved items requires a model to sample two or more negatives differing in preferences.
In a basic form, let $p_1$ and $p_2$ represent two different sampling distributions for negatives.
Given a user $u$ and its positive $i$,
we sample a negative $j$ from $p_1$ and $k$ from $p_2$ where $i \neq j \neq k$.
Since the positive $i$ is the most preferred item to the user $u$,
a transitive relation $\hat{s}_{ui} > \hat{s}_{uj} > \hat{s}_{uk}$ holds true when $j$ is more preferred than $k$.
Strict transitivity then corresponds to a scheme where any $j$ from $p_1$ is guaranteed to be more preferred than any $k$ from $p_2$.
However, when the sampled negative $j$ is actually less preferable than $k$, the relation $\hat{s}_{ui} > \hat{s}_{uj} > \hat{s}_{uk}$ is violated.
We refer this scheme as weak transitivity which allows such occasional violations.

The concept of transitivity with multiple negatives is straightforward as well.
Instead of sampling a single negative from each distribution, we sample a set of negatives $\mathcal{N}_j = \{j_1, \ldots, j_n\}$ from $p_1$ and $\mathcal{N}_k = \{k_1, \ldots, k_n\}$ from $p_2$.
Consequently, with $\hat{s}_{\mathcal{N}_j} = \{\hat{s}_{uj_1}, \ldots, \hat{s}_{uj_n}\}$ and $\hat{s}_{\mathcal{N}_k} = \{\hat{s}_{uk_1}, \ldots, \hat{s}_{uk_n}\}$,
a transitive relation is modified as $\hat{s}_{ui} > \max(\hat{s}_{\mathcal{N}_j}) > \min(\hat{s}_{\mathcal{N}_j}) > \max(\hat{s}_{\mathcal{N}_k})$ with the same criteria for weak and strict cases.

\subsection{Extensions with Weak Transitivity}

We propose novel extensions of original training objectives that resolves the limitation of binary label assignments by incorporating the derived transitive relation to the loss formulation.
We first introduce two sampling schemes by utilizing an item popularity distribution $p_{\textrm{pop}}$ and a uniform distribution $p_{\textrm{unif}}$:
\begin{align}
    \mathcal{D}'_s(\textrm{pop}) &= \{(u,i,j,k) \mid j \sim p_{\textrm{pop}}, k \sim p_{\textrm{unif}}\},
    \label{eqn:wts_pop}\\
    \mathcal{D}'_s(\textrm{niche}) &= \{(u,i,j,k) \mid j \sim p_{\textrm{unif}}, k \sim p_{\textrm{pop}}\},
    \label{eqn:wts_niche}
\end{align}
where $j$ is the more preferred negative and $k$ is the less preferred one. 
The mini-batch $\mathcal{D}'_s(\textrm{pop})$ assumes that a user generally prefers popular items whereas $\mathcal{D}'_s(\textrm{niche})$ regards niche items more favored.
Our extension of the BPR objective with transitivity is then given by the following:
\begin{equation}
    \mathcal{L}_{\textrm{TransBPR}}
    = -
    \!\sum_{(u,i,j,k)\! \in \mathcal{D}'_s}
    \underbrace{
    \log \sigma(\hat{s}_{ui} - \hat{s}_{uj})}_{\textrm{original}} + 
    \gamma\underbrace{
    \log \sigma(\hat{s}_{uj} - \hat{s}_{uk})}_{\textrm{preference}},
    \label{eqn:transbpr_loss}
\end{equation}
where $\gamma$ is a balancing coefficient for two terms.
In essence, the preference term encourages the score of $j$ to be higher than that of $k$ while the original term assures it to be smaller than that of $i$.
As a consequence, our objective explicitly imposes the transitive relation $\hat{s}_{ui} > \hat{s}_{uj} > \hat{s}_{uk}$ to recommendation scores. 
Combining the proposed formulation with the previously introduced sampling schemes, we obtain two distinct training objectives as follows:
\begin{align}
    \textrm{TransBPR}_{\textrm{pop}} &=
    \mathcal{L}_{\textrm{TransBPR}}(\mathcal{D}'_s(\textrm{pop}); \theta),\\
    \textrm{TransBPR}_{\textrm{niche}} &=
    \mathcal{L}_{\textrm{TransBPR}}(\mathcal{D}'_s(\textrm{niche}); \theta),
\end{align}
where $\theta$ is learnable parameters of the model.
Similarly, the extension of the BCE function is formulated as below:
\begin{align}
    \mathcal{L}_{\textrm{TransBCE}} = & -
    \sum_{(u,i,j,k) \in \mathcal{D}'_s}
    \Bigl[ \log(\sigma(\hat{s}_{ui})) + 
    \log(1 - \sigma(\hat{s}_{uj}))
    \nonumber\\
    & + \gamma \Bigl(
    \log(\sigma(\hat{s}_{uj})) + 
    \log(1 - \sigma(\hat{s}_{uk})) \Bigl)\Bigl].
    \label{eqn:transbce_loss}
\end{align}
Naturally, $\textrm{TransBCE}_{\textrm{pop}}$ and $\textrm{TransBCE}_{\textrm{niche}}$ are two consequent training schemes with $\mathcal{D}'_s(\textrm{pop})$ and $\mathcal{D}'_s(\textrm{niche})$, respectively.
For the setwise loss function, we sample a set of items $\mathcal{N}_j$ and $\mathcal{N}_k$ from each distribution instead of sampling a single item $j$ or $k$:
\begin{align}
    \mathcal{D}'_m(\textrm{pop}) &= \{(u,i,\mathcal{N}_j,\mathcal{N}_k) \mid \mathcal{N}_j \sim p_{\textrm{pop}}, \mathcal{N}_k \sim p_{\textrm{unif}}\},
    \label{eqn:wtm_pop}\\
    \mathcal{D}'_m(\textrm{niche}) &= \{(u,i,\mathcal{N}_j,\mathcal{N}_k) \mid \mathcal{N}_j \sim p_{\textrm{unif}}, \mathcal{N}_k \sim p_{\textrm{pop}}\}.
    \label{eqn:wtm_niche}
\end{align}
A corresponding extension for the SSM loss function is given by:
\begin{align}
    \mathcal{L}_{\textrm{TransSSM}} = & -
    \!\sum_{(u,i,\mathcal{N}_j,\mathcal{N}_k) \in \mathcal{D}'_m}\!
    \biggl[
    \log\frac{\exp(\hat{s}_{ui})}
    {\exp(\hat{s}_{ui}) + 
    \sum_{j \in \mathcal{N}_j}\exp(\hat{s}_{uj})}
    \nonumber\\
    &+ \gamma\frac{1}{|\mathcal{N}_j|}
    \!\sum_{j \in \mathcal{N}_j}\!
    \log\frac{\exp(\hat{s}_{uj})}
    {\exp(\hat{s}_{uj}) + 
    \sum_{k \in \mathcal{N}_k}\exp(\hat{s}_{uk})}
    \biggl].
    \label{eqn:transssm_loss}
\end{align}
As similar to previous extensions, $\textrm{TransSSM}_{\textrm{pop}}$ and $\textrm{TransSSM}_{\textrm{niche}}$ employs either $\mathcal{D}'_m(\textrm{pop})$ or $\mathcal{D}'_m(\textrm{niche})$ for negative sampling.

One thing to note is that our introduced sampling strategies $\mathcal{D}'_s$ and $\mathcal{D}'_m$ are both weak transitive.
Nonetheless, modifying them to a strict setting can be readily accomplished if we can quantify the preference of each item.
For instance, we can reformulate the mini-batch construction for pairwise and pointwise objectives as:
\begin{align}
    \mathcal{D}^\dag_s(\textrm{pop}) &= \{(u,i,j,k) \mid f(j) > f(k), j \sim p_{\textrm{pop}}, k \sim p_{\textrm{unif}}\},
    \label{eqn:sts_pop}\\
    \mathcal{D}^\dag_s(\textrm{niche}) &= \{(u,i,j,k) \mid f(j) < f(k), j \sim p_{\textrm{unif}}, k \sim p_{\textrm{pop}}\},
    \label{eqn:sts_niche}
\end{align}
where $f(\cdot)$ denotes a function that measures the popularity of an item.
However, we argue that such strict transitivity rather hurts the quality of the resulting recommendation policy.

\begin{table}[t]
    \caption{Comparison of our methods with strict transitivity and weak transitivity in the Amazon Beauty dataset.}
    \label{tab:transitivity_ablation}
    \setlength{\tabcolsep}{12pt}
    \begin{center}
    \begin{tabular}{lxxy}
    \toprule
    \rowcolor{white}
    \multicolumn{2}{l}{\textbf{Method}} 
    & \textbf{HR} & \textbf{NDCG} \\
    \midrule
    \rowcolor{white}
    & Strict & 0.0508 & 0.0229 \\
    \multirow{-2}{*}{$\textrm{TransBPR}_{\textrm{pop}}$}
    & Weak & 0.0674 & 0.0293 \\
    \midrule
    \rowcolor{white}
    & Strict & 0.0571 & 0.0252 \\
    \multirow{-2}{*}{$\textrm{TransBPR}_{\textrm{niche}}$}
    & Weak & 0.0651 & 0.0289 \\
    \midrule
    \rowcolor{white}
    & Strict & 0.0697 & 0.0311 \\
    \multirow{-2}{*}{$\textrm{TransBCE}_{\textrm{pop}}$}
    & Weak & 0.0730 & 0.0324 \\
    \midrule
    \rowcolor{white}
    & Strict & 0.0635 & 0.0282 \\
    \multirow{-2}{*}{$\textrm{TransBCE}_{\textrm{niche}}$}
    & Weak & 0.0720 & 0.0322 \\
    \midrule
    \rowcolor{white}
    & Strict & 0.0631 & 0.0309 \\
    \multirow{-2}{*}{$\textrm{TransSSM}_{\textrm{pop}}$}
    & Weak & 0.0843 & 0.0395 \\
    \midrule
    \rowcolor{white}
    & Strict & 0.0700 & 0.0346 \\
    \multirow{-2}{*}{$\textrm{TransSSM}_{\textrm{niche}}$}
    & Weak & 0.0774 & 0.0370 \\
    \bottomrule
    \end{tabular}
    \end{center}
\end{table}

\section{Experiments}
\label{sec:experiments}

In this section, we conduct experiments to compare our proposed extensions to original objectives and their variants.

\subsection{Experimental Setup}

\paragraph{Datasets}
We employ public sequential recommendation tasks from different domains:
\emph{Beauty}, \emph{Toys}, and \emph{Sports},
which are product review datasets introduced by Amazon.com~\cite{mcauley2015image},
and \emph{Yelp}, which is a widely tested business recommendation dataset. 
Detailed statistics of preprocessed datasets are reported in Table~\ref{tab:datasets}.

\paragraph{Evaluation Settings and Metrics}
For dataset partitioning, we adopt the conventional \emph{leave-one-out} strategy~\cite{kang2018self, sun2019bert4rec} to assure the quality of the trained recommender system.
Then, we recommend $K$ items with the highest recommendation scores from the entire item pool.
For evaluation, we adopt two common top-$K$ metrics, HR@$K$ and NDCG@$K$ with $K$ fixed to 10.

\paragraph{Baselines}
We fix a model architecture to \emph{SASRec}~\cite{kang2018self}
and switch only a training objective. 
For baselines, we compare our method against \emph{BPR}~\cite{Rendle2009BPR},
\emph{GBPR}~\cite{pan2013gbpr},
\emph{BCE}~\cite{kang2018self},
\emph{SSM}~\cite{wu2024effectiveness},
\emph{InfoNCE}~\cite{oord2018representation},
\emph{BPR-DNS}~\cite{zhang2013optimizing},
and \emph{gBCE}~\cite{petrov2023gsasrec}.
Here, $\textrm{gBCE}_{1}$ and $\textrm{gBCE}_{\mathcal{N}}$ denote gBCE with a single negative and multiple negatives, respectively.

\paragraph{Hyperparameters}
For all objectives, We train with a fixed batch size of 256, a learning rate of 0.0003, and a maximum sequence length of 50.
The SASRec model is with 2 layers and 1 attention head with an embedding dimension of 256. 
For setwise objectives, we sample 100 negatives in total.
Our proposed TransSSM sets cardinality of $\mathcal{N}_j$ and $\mathcal{N}_k$ to 50 such that they sum to 100.
A balancing coefficient $\gamma$ is selected from $\{0.5, 1.0, 1.5\}$.

\subsection{Performance Comparison}

Table~\ref{tab:main_results} summarizes the performance of models trained with different optimization objectives. 
In general, our proposed objectives outperform three original objectives and their notable variants in both metrics within all benchmarks.
Particularly, we recognize $\textrm{TransSSM}_{\textrm{pop}}$ obtain the highest performance among all baselines in all datasets.
Such results demonstrate that inducing preference order through weak transitivity consistently improves the performance regardless of base loss functions.
Typically, we observe our extensions with popularity preference (e.g., $\textrm{TransSSM}_{\textrm{pop}}$) achieve commonly improved metrics compared to niche preference (e.g., $\textrm{TransSSM}_{\textrm{niche}}$).
Hence, exploiting popularity as the preference indicator is particularly effective regardless of datasets.
Though, we find recommendations with niche preference performs on par with best-performing baselines or often even better.

\begin{figure}[t]
    \centering
    \subfigure[$\log \sigma(\hat{x}_{uij})$ of $\textrm{TransBPR}_{\textrm{pop}}$]{
        \includegraphics[width=0.46\columnwidth]{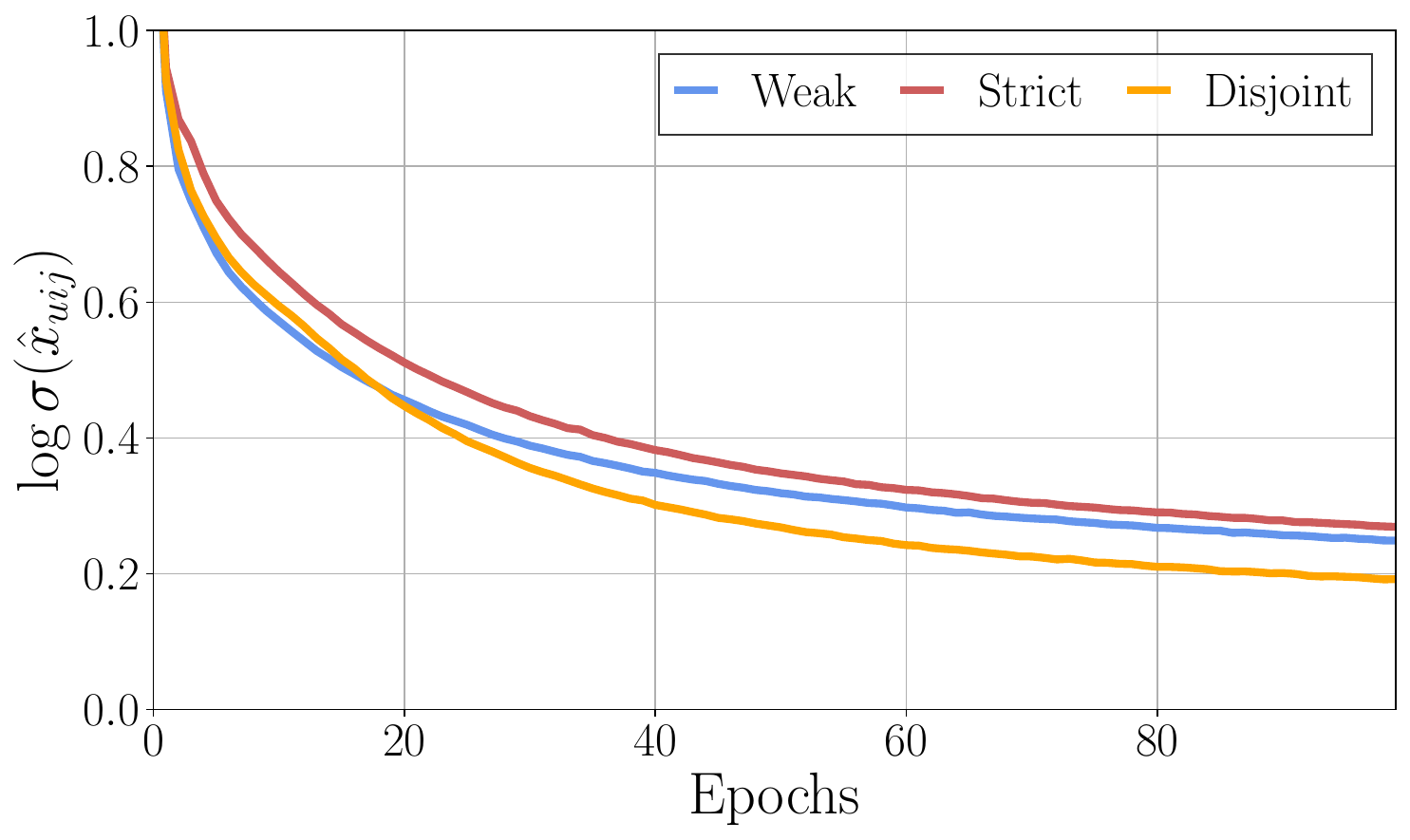}
        \label{fig:bpr_pop_left}
    }
    \subfigure[$\log \sigma(\hat{x}_{ujk})$ of $\textrm{TransBPR}_{\textrm{pop}}$]{
        \includegraphics[width=0.46\columnwidth]{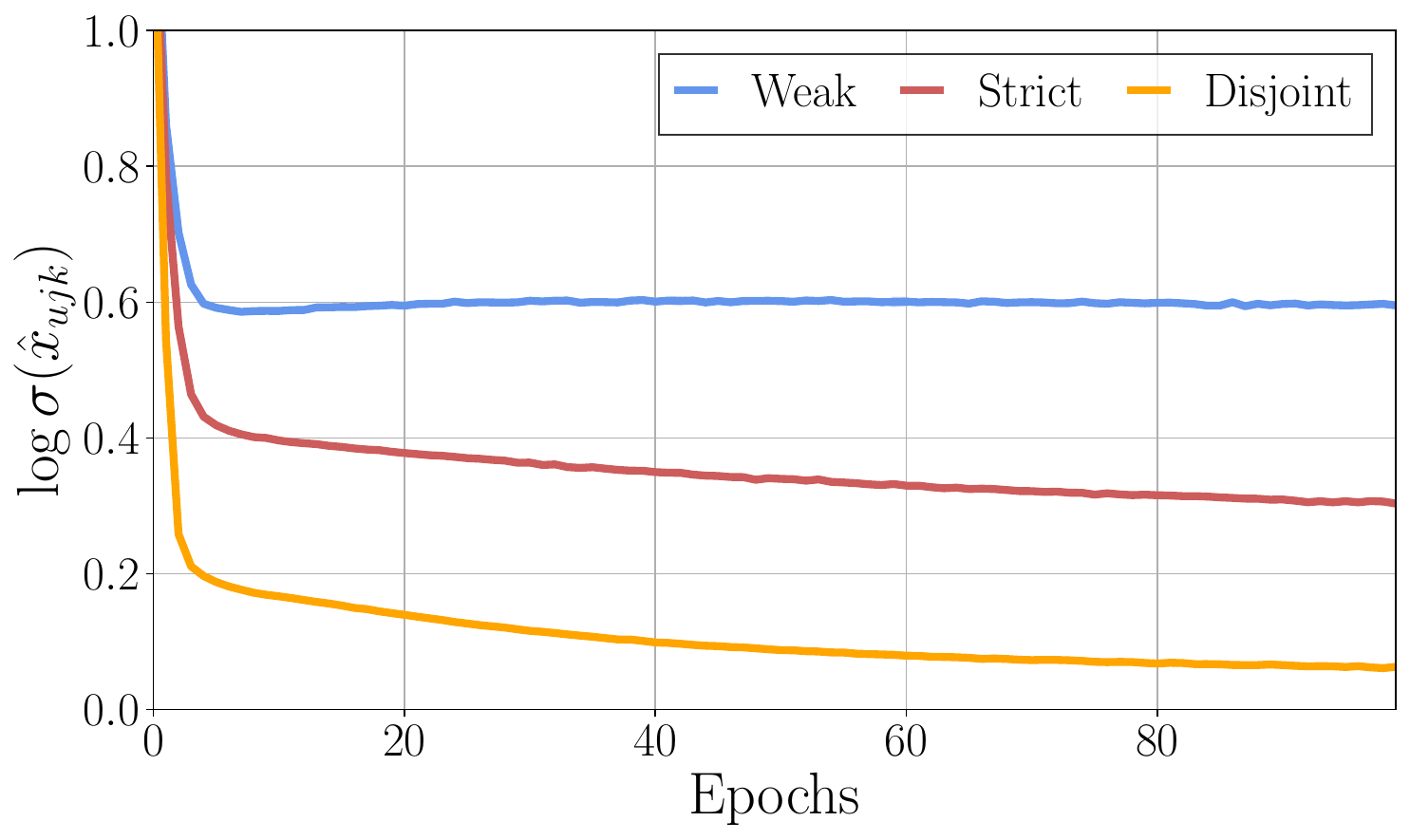}
        \label{fig:bpr_pop_right}
    }
    \subfigure[$\log \sigma(\hat{x}_{uij})$ of $\textrm{TransBPR}_{\textrm{niche}}$]{
        \includegraphics[width=0.46\columnwidth]{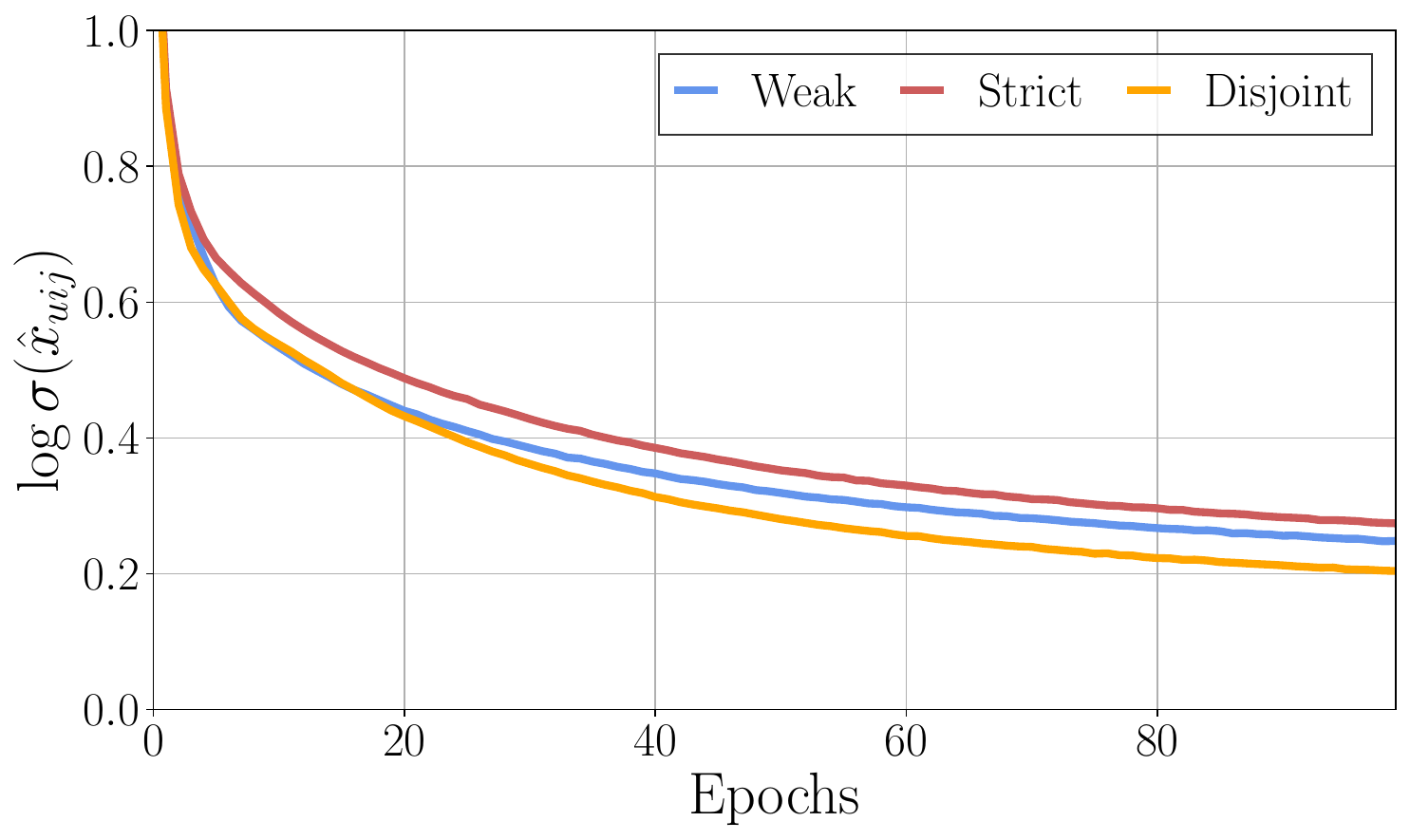}
        \label{fig:bpr_niche_left}
    }
    \subfigure[$\log \sigma(\hat{x}_{ujk})$ of $\textrm{TransBPR}_{\textrm{niche}}$]{
        \includegraphics[width=0.46\columnwidth]{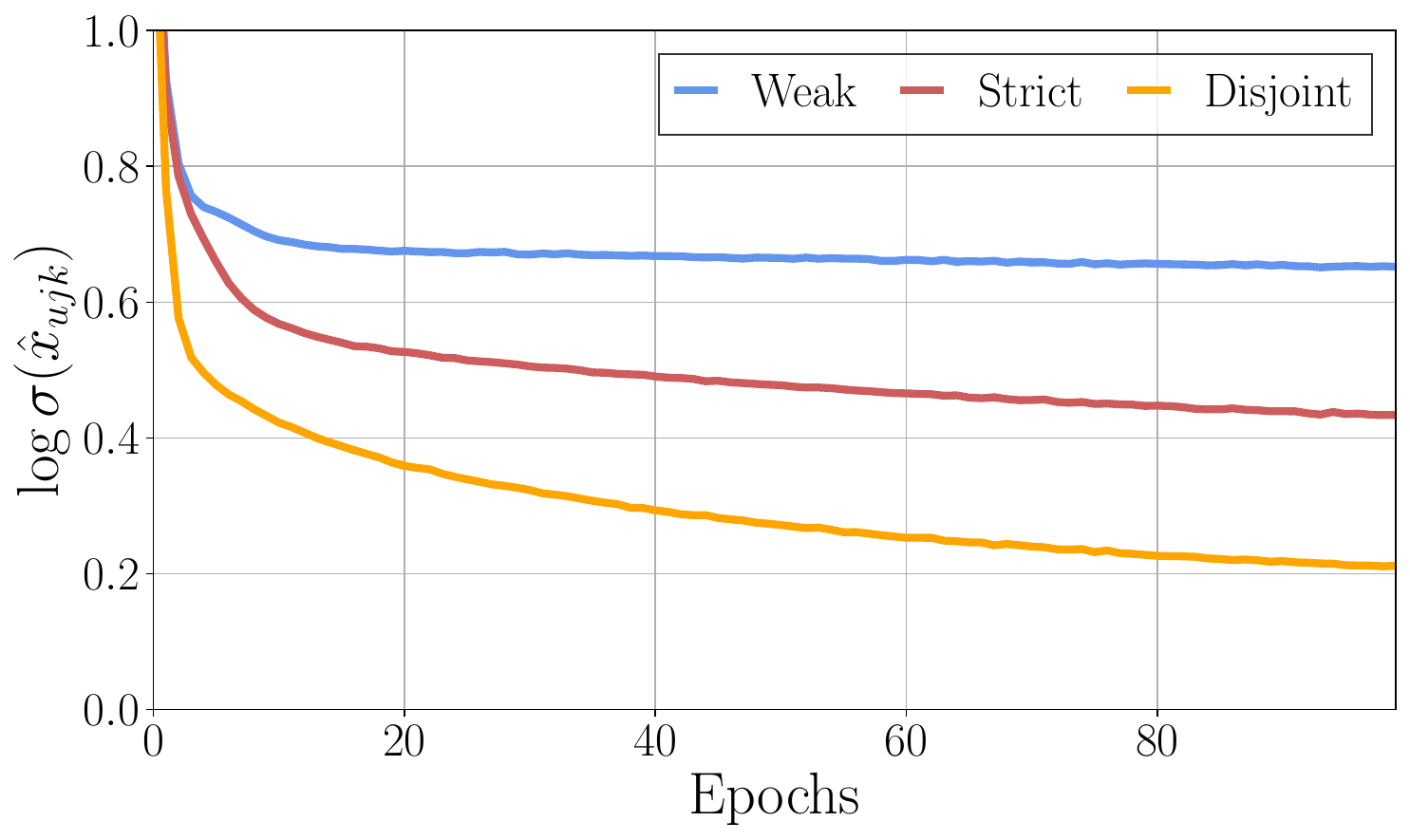}
        \label{fig:bpr_niche_right}
    }
    \caption{Average values of each loss term in TransBPR with Weak (blue), Strict (red), and Disjoint (yellow) transitivity.}
    \Description{Average values of each loss component}
    \label{fig:loss_analysis}
\end{figure}

\subsection{Transitivity Analysis}

As seen in Table~\ref{tab:transitivity_ablation},
we observe comparatively higher performance with weak transitivity.
We hypothesize that the number of uninformative (i.e., easy) negatives~\cite{rendle2014improving, wang2018incorporating} increases more within the strict transitivity as training proceeds.
To validate the claim, we report how average value of each term in the BPR extension,
$\log \sigma(\hat{x}_{uij})$ and $\log \sigma(\hat{x}_{ujk})$, changes throughout the training with different schemes.
Here, we compare three classes of transitivity:
\emph{Weak}, i.e., \eqref{eqn:wts_pop} and \eqref{eqn:wts_niche},
\emph{Strict}, i.e., \eqref{eqn:sts_pop} and \eqref{eqn:sts_niche},
and \emph{Disjoint}, i.e., a specific case of Strict that sets $50\%$ of total items with high preference as a support for $p_{\textrm{pop}}$ and the rest for $p_{\textrm{niche}}$.
Results in Figure~\ref{fig:loss_analysis} illustrate more steep decline of the preference term, $\log \sigma(\hat{x}_{ujk})$, as the transitivity becomes more strict.
Thus, weak transitivity provides the most informative gradients, increasing the training effectiveness.

\section{Conclusion}

In this work, we identified the common and crucial disadvantage of binary label assignments within conventional recommendation objectives.
To overcome this limitation, we have proposed novel extensions that directly exploit the preference differences of unobserved items.
Through extensive experiments, we demonstrated the effectiveness of our method with the thorough analysis.
As future work,
the uniformity of normalized embeddings~\cite{ChungH2023ssltp} can be applied to our framework to improve recommendation performance more.

\bibliographystyle{ACM-Reference-Format}
\bibliography{sample-base}

\end{document}